\documentclass[10pt,letterpaper]{article}
\usepackage[top=0.85in,left=2.75in,footskip=0.75in]{geometry}

\usepackage{amsmath,amssymb}

\usepackage{changepage}

\usepackage[utf8x]{inputenc}

\usepackage{textcomp,marvosym}

\usepackage{cite}
\usepackage{enumitem}

\usepackage{nameref,hyperref}

\usepackage{microtype}
\DisableLigatures[f]{encoding = *, family = * }

\usepackage[table]{xcolor}

\usepackage{array}
\usepackage{multicol}
\usepackage{multirow}
\usepackage{booktabs}

\newcolumntype{+}{!{\vrule width 2pt}}

\newlength\savedwidth

\makeatletter
\newcommand{\justified}{%
  \rightskip\z@skip%
  \leftskip\z@skip}
\makeatother

\raggedright
\usepackage[aboveskip=1pt,labelfont=bf,labelsep=period,justification=raggedright,singlelinecheck=off]{caption}

\makeatletter
\renewcommand{\@biblabel}[1]{\quad#1.}
\makeatother

\usepackage{lastpage,fancyhdr,graphicx}
\usepackage{epstopdf}
\fancyheadoffset[L]{2.25in}
\fancyfootoffset[L]{2.25in}
\begin{document}
\vspace*{0.2in}

\begin{flushleft}
{\Large
\textbf\newline{``When they say weed causes depression, but it’s your fav antidepressant": Knowledge-aware Attention Framework for Relationship Extraction} 
}
\newline
\\
Shweta Yadav\textsuperscript{1\ddag},
Usha Lokala\textsuperscript{2},
Raminta Daniulaityte\textsuperscript{3},
Krishnaprasad Thirunarayan\textsuperscript{1},
Francois Lamy\textsuperscript{4},
Amit Sheth\textsuperscript{2}
\\
\bigskip
\textbf{1} Wright State University, Dayton, Ohio, USA
\\
\textbf{2} University of South Carolina, Columbia, South Carolina, USA
\\
\textbf{3} Arizona State University, Tempe, Arizona, USA
\\
\textbf{4} Mahidol University, Thailand
\\
\bigskip

%
%


\textcurrency Current Address: Wright State University, Dayton, Ohio, USA 



* shweta@knoesis.org

\end{flushleft}
\section*{Abstract}

With the increasing legalization of medical and recreational use of cannabis, more research is needed to understand the association between depression and consumer behavior related to cannabis consumption.
Big social media data has potential to provide   deeper insights about these associations to public health analysts.
In this interdisciplinary study, we  demonstrate the value of incorporating domain-specific knowledge in the learning process to identify the relationships between cannabis use and depression. 
We develop an end-to-end knowledge infused deep learning framework (Gated-K-BERT) that leverages the pre-trained BERT language representation model and domain-specific declarative knowledge source (Drug Abuse Ontology (DAO)) to jointly extract entities and their relationship using gated fusion sharing mechanism. 
Our model is further tailored to provide more focus to the entities mention in the sentence through entity-position aware attention layer, where ontology is used to locate the target entities position. 
Experimental results show that inclusion of the knowledge-aware attentive representation in association with BERT can extract the cannabis-depression relationship with better coverage in comparison to the state-of-the-art relation extractor.



\section{Introduction}
Over 30 states in the US have now passed laws legalizing comprehensive medical cannabis programs. Since 2012, 11 states have legalized the recreational use of cannabis \cite{Hanson2014-jh}. 
\cite{Room2014-fk, Volkow2014-jl} discuss epidemiological monitoring of therapeutic uses of cannabis products needed to assess the impact of policy changes, and identifying emerging issues and trends. Although the prevalence of depression in the US population, notably among young adults \cite{Weinberger2018-bv} has increased, and a variety of pharmacological treatments are available, a large proportion of individuals with depression delay seeking treatment or avoid it altogether \cite{Young2008-jg}. Current medical cannabis policies across the US do not include depression as a medical qualifying condition related to medical cannabis use \cite{Bridgeman2017-er}. However, emerging research indicates that coping with depression is often reported as an important reason for cannabis use \cite{Lankenau2018-rh}. 
While researchers have found that the cannabis has general therapeutic benefits \cite{Keyes2016-wd}, more research is needed to understand the evolving trends  and depressive behaviors related to cannabis consumption. \\
\indent In this context, social media platforms play an  important role in uncovering experiences of individuals and their health-related knowledge \cite{Corazza2013-qf, Burns2014-ic}. Although  user generated content area a rich source of unsolicited and unfiltered self-disclosures of attitudes and practices related to cannabis use the relationship between cannabis and depression remains ambiguous  \cite{Cavazos-Rehg2018-zc, Daniulaityte2017-fc, Lamy2018-ou}.
We formulate this problem as the extraction of relationship between cannabis use and depression in terms of four possible relationships namely: Reason, Effect, Addiction, and Ambiguous (c.f. Table \ref{tab:table1}). We have identified from the literature that the cannabis use can be a reason for depression or an effect of depression.
\begin{table}[h!]
  \begin{center}
  \resizebox{\linewidth}{!}{
\begin{tabular}{l|l}
 \hline
      \textbf{Relationship} & \textbf{{Tweet}}\\
      \hline
      {Reason}  
      & ``-Not saying im cured, but i feel less \textcolor{red}{depressed} lately, could be my \textcolor{blue}{CBD oil} supplement." \\ 
      \hline
      Effect  & ``-People will \textcolor{blue}{smoke weed} and be on antidepressants. It's a clash!Weed is what is making you \textcolor{red}{depressed}." \\
      \hline
      Addiction & ``-The lack of \textcolor{blue}{weed} in my life is \textcolor{red}{depression} as hell." \\ 
      \hline
      Ambiguous  & ``-People with an aversion to \textcolor{blue}{weed} heavily are like intentionally \textcolor{red}{depressed}." \\
      \hline
    \end{tabular}
    }
     \caption{Cannabis-Depression Tweets and their relationships. Here the text in the \textcolor{blue}{blue} and \textcolor{red}{red} represents the cannabis and depression entities respectively.}
    \label{tab:table1}
  \end{center}
\end{table}
Extracting relationships between any \textit{concepts/slang-terms/synonyms/street-names} related to `\textit{cannabis}', and similarly those related to `\textit{depression}', requires a domain ontology. Here, we use Drug Abuse Ontology (DAO) \cite{Cameron2013-qq,noauthor_undated-xp} which is a domain-specific hierarchical framework containing 315 entities (814 instances) and 31 relations defining drug-abuse and mental-health disorder concepts. The ontology has been utilized in analyzing web-forum content related to buprenorphine, cannabis, a synthetic cannabinoid, and opioid-related data \cite{Lokala2018-dm, cameron2013predose,kumar2020edarkfind}. DAO was expanded using DSM-5 categories covering mental health and applied in this work for improving mental health associations with cannabis on Twitter \cite{Gaur2018-xv}.\\
\indent For entity and relationship extraction (RE) task, previous approaches generally adopt deep learning models \cite{yadav2017entity,srivastava2016recurrent,yadav2016deep}, in particular, Convolutional Neural Network (CNN) \cite{lin2016neural,ekbal2016deep} and Bi-directional Long Short Term Memory (Bi-LSTM) \cite{lee2019semantic,yadav2020exploring,yadav2019unified} networks. However, Bi-LSTM/CNN model does not generalize well and performs poorly in limited supervision scenarios.
Recently, several pre-trained language representation models have  significantly advanced the state-of-the-art in various NLP tasks \cite{maillard2019jointly,akbik2019pooled}. BERT \cite{devlin-etal-2019-bert} is one of the powerful language representation models that has the ability to make predictions that go beyond the natural sentence boundaries \cite{lin2019bert}.
Unlike CNN/LSTM model, language models benefit from the abundant knowledge from pre-training using self-supervision and have strong feature extraction capability. So we exploit the representation from BERT and CNN to achieve best of both the representations using novel gating fusion mechanism. Further, we tailored our model to capture the entities position information (using DAO knowledge) which is crucial for the RE as established in the prior research \cite{zhou2018position,he2018see}.\\
\indent We propose an end-to-end knowledge-infused deep learning framework (named, \textit{\textbf{Gated-K-BERT}}) based on widely adopted BERT language representation model and domain-specific DAO ontology to extract entities and their relationship. The proposed model has three modules: 
\textit{\textbf{(1) Entity Locator}}, which utilizes the DAO ontology to map the input word sequence to the entities mention in the ontology by computing the edit distance between the entity names (obtained from the DAO) and every n-gram token of the input sentence. 
\textit{\textbf{(2) Entity Position-aware Module}}, exploits the DAO to explicitly integrate the knowledge of entities in the model. This is done by encoding position sequence relative to the entities. Further, we make the attention layers aware of the positions of all entities in the sentence.
\textit{\textbf{(3) Encoding Module}}, jointly leverages the distributed representation obtained from BERT and entity position-aware module using the shared gated fusion layer
to learn the contextualized syntactic and semantic information which are complimentary to each other. \\
\textbf{Contributions}:\\
\textbf{(1)} In collaboration with domain experts, we introduce an annotation scheme to label the relationships between cannabis and depression entities to generate a gold standard cannabis-depression relationship dataset extracted using Twitter. 
\\
\textbf{(2)} We propose an end-to-end knowledge-infused neural model to extract cannabis/depression entities and predict the relationship between those entities. We exploited domain-specific DAO ontology which provides better coverage in entity extraction. We further augment the BERT model into knowledge-aware framework using gated fusion layer to learn the joint feature representation.\\
\textbf{(3)} We explored entity position-aware attention in the task to
jointly leverages the distributed representation of word position relative to cannabis/depression mention and the attention mechanism. \\
\textbf{(4)} We evaluated our proposed model on real-world social media dataset. The experimental results shows that our model outperforms the state-of-the-art relation extraction techniques. We further analyzed that enhancing neural attention with entity position knowledge improves the performance of the model to predict the correct relationship between cannabis and depression over vanilla attention mechanism.
\section{Related Work}
Based on the techniques, recent existing works can be broadly categorized into the following:
\begin{enumerate}
\item{\textbf{Deep Learning (DL) Framework:}} Several DL approaches primarily based on CNN \cite{liu2013convolution} and LSTM \cite{miwa2016end,yadav2019feature,ningthoujam2019relation,yadav2018feature1} techniques has been proposed for RE. A study by \cite{Liang2017-hy} develops a hybrid deep neural network model using Bi-Directional Gated Recurrent Neural Network (Bi-GRU), CNN, GRU, and Highway connection for classifying relations in SemEval 2010 and KBP-SF48 dataset. 
\cite{pmlr-v97-wu19e} exploited the dependency tree by utilizing Graph Convolutional Neural Network (GCN) to capture rich structural information that has been demonstrated for the RE task. \cite{guo-etal-2019-attention} advanced the previous methods based on GCN by guiding the network through the attention mechanism. Another prominent work by \cite{bekoulis-etal-2018-adversarial} explores the adversarial learning to jointly extract entities and their relationship. To further enhance the performance of the DL models, various techniques \cite{choi2018extraction,peng2017deep} has also exploited latent features in particular the entity position information in the DL framework. 
\item{\textbf{Pre-trained Language Representation Model:}} Models such as BERT, BioBERT \cite{lee2019biobert}, SciBERT \cite{beltagy2019scibert}, and XLNet \cite{yang2019xlnet} has shown the state-of-the-art performance on RE task. \cite{shi2019simple} adapted the BERT for the relation extraction and semantic role labeling task. \cite{xue2019fine} modified the BERT framework by constructing task-specific MASK that control the attention in last layers of the BERT. 
\cite{wang2019extracting} also modified the original BERT architecture by introducing a structured prediction layer that is able to predict the multiple relations in one pass and make attention layers aware of the entities position.
\item{\textbf{Knowledge-base Framework:}} Study by \cite{Chan2010-pg} saw the importance of external knowledge in improving the relation extraction from sentences. The study utilizes the parent-child relationships in Wikipedia and word cluster over unlabeled data into a global inference procedure using Integer Linear Programming (ILP). Experiments conducted on ACE-2004 dataset show that the use of background knowledge improved F-measure by $3.9$\%.
A study by \cite{Wen2018-en} uses the attention model to traverse a medical knowledge graph for entity pairs which assist in precise relation extraction. \cite{distiawan2019neural} jointly learn the word and entity embedding (obtained through the TransE) using the anchor context model to extract the relationship and the entities. Some of the other prominent work utilizing knowledge graph for relation extraction are \cite{li2019dual,zhou2019knowledge,li2019improving}.
\end{enumerate}
\section{Resource creation and annotation scheme} \label{dataset}
We created a corpus by collecting the tweets from the time period of January 2017 to February 2019 using Twitter data processing, filtering, and aggregation framework available through the `Twitris' \cite{} which has been configured to collect tweets with relevant keywords and adapted to perform appropriate analysis. From the available corpus of over 100 million relevant tweets collected so far, we further filtered tweets using DAO based on Cannabis and Depression entities and their respective instances specifically defined by domain experts (substance use epidemiologist)  for this context. From that filtered corpus, a sample of around 11000 tweets was sent for expert annotation to a team of 3 substance use epidemiologist co-authors who have vast experience in Interventions, Treatment and Addictions Research. Further processing was done on this corpus based on the tweets lacking one of the key concepts related to cannabis/depression and 5885 tweets were annotated finally. The annotation scheme is based on the following coding:
\begin{figure*}[t]
\centering
\includegraphics[width=\linewidth]{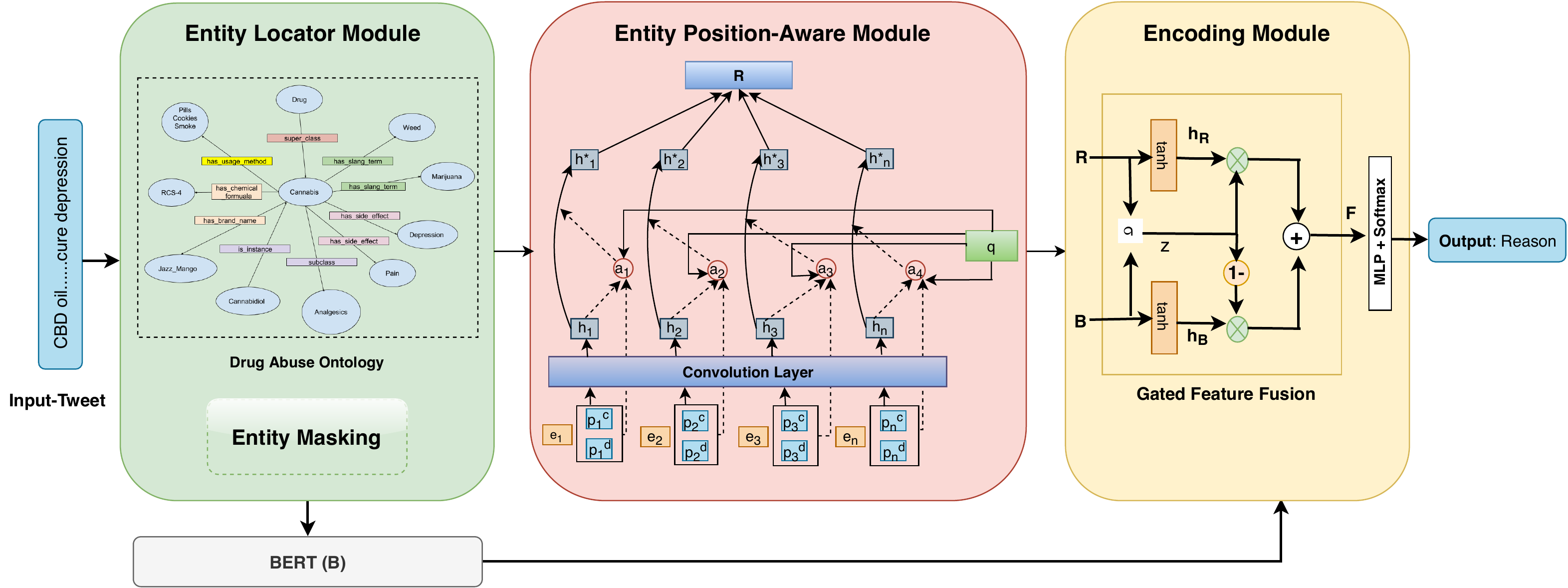}
\label{Figure-1}
\caption{Proposed model architecture for cannabis-depression relation extraction. Input to the model is the tweet and output is the relationship between cannabis and depression entity.}
\end{figure*}
\begin{enumerate}
    \item \textbf{Reason:} Cannabis is used to help/treat/cure depression. 
    \item \textbf{Effect:} Cannabis causes depression or makes symptoms worse.
    \item \textbf{Addiction:} Lack of access to cannabis leads to depression, showing potential symptom of addiction.
    \item \textbf{Ambiguous:} Implies other types or relationships, or too ambiguous/unclear to interpret.
\end{enumerate}

\begin{table}
\footnotesize
\centering
\begin{tabular}{|c|c|c|c|}
         \toprule
         & \textbf{B} & \textbf{C} & \textbf{D} \\ \midrule
         \textbf{A} & \textbf{0.83} & 0.79 & 0.75 \\ 
         \textbf{B} & - & 0.75 & \textbf{0.86} \\
         \textbf{C} & - & - & 0.80 \\ 
         \bottomrule
    \end{tabular}
\caption{ Pairwise average annotator agreement using Cohens Kappa between 4 annotators over 3 cycles.}
\label{tab:kappa}
\end{table}
The category ``Addiction" is an intermediate between the first two as it indicates that feelings of depression would be resolved if one had access to cannabis (which relates to category 1) and suggests that the  presence of cannabis withdrawal symptoms (which relates to category 2). Due to the brevity and ambiguity of information provided in the tweet content, the team decided to classify such cases as a separate category. 

The sub-samples of tweets were coded independently by each coder and an inter-coder agreement was calculated. The team went through 3 iterations of coding, assessing and discussing, disagreement, and improving coding rules until an acceptable level of agreement was reached among coders (Cohen’s kappa of 0.80,(\textit{c.f.}\ref{tab:kappa})) \cite{McHugh2012-fo}.  Tweets that were coded differently by two primary coders were reviewed by a third coder to resolve the disagreement. This yielded a dataset containing 5885 tweets out of which (1) 3243 tweets are annotated as `Reason' (2) 707 tweets are annotated as `Effect'. (3) 158 tweets are annotated as `Addiction' (4) 1777 tweets are annotated as `Ambiguous'. The mean tweet text length is 148 tokens (median 74). 

The university institutional review board (IRB) approved the study under Human Subjects Research Exemption 4 because it is limited to publicly available tweets. To protect anonymity, cited tweet content was modified slightly. We note that this dataset has some (inevitable) limitations: (i) the method only captures a sub-population of cannabis-depression related tweets in eDrugTrends campaign (i.e. those with terms defined in ontology), (ii) Tweets collected may not be a representative sample of the population as a whole, and (iii) there is no way to verify whether the tweets with self-reported cannabis related depression or cannabis related relief from depression are truthful. Our substance use epidemiologist co-authors established the validity and relevance of the final set of annotated tweets. 

\textbf{Ethics:} Our project involves analysis of Twitter data that is publicly available and that has been anonymized. It does not involve any direct interaction with any individuals or their personally identifiable data. So our work does not meet the Federal definition for human subjects research, specifically, “a systematic investigation designed to contribute to generalizable knowledge” and “research involving interaction with the individual or obtains personally identifiable private information about an individual”. Thus, this study was reviewed by the Wright State University IRB and received an exemption determination.
\section{Our Proposed Approach}
In this study, a knowledge-infused RE framework, Gated Knowledge BERT (Gated-K-BERT) is used to identify relations between entities `\textit{cannabis}' and `\textit{depression}' in a tweet. 
Our framework consists of three components discussed as follows:
\subsection{Entity Locator Module}
Let $S$ be an input tweet containing the $n$ words $\{w_1, w_2, \ldots, w_n\}$. Extracting relationships between any \textit{concepts/slang-terms/synonyms/street-names} related to `\textit{cannabis}' and similarly those related to `\textit{depression}' require heavy dependency on the domain knowledge model. 
We used domain-specific DAO to map entities in a tweet to their parent concepts in the ontology by computing the edit distance between the entity names (obtained from  the  DAO)  and  every  n-gram  token  of  the input  sentence. Since, DAO provides much better coverage on the entities, it is assume that entity name will be mention in the sentence. \\
\indent Later, we perform masking on the extracted entities. The reason for masking is to explicitly provide the model with the entity information and also prevent a model from overfitting its predictions to specific entities.
 For instance, entities related to cannabis in a tweet are masked by `\textit{$<$cannabis$>$}'. Similarly, entities related to depression are masked with `\textit{$<$depression$>$}'. 
 By this, we obtain a cannabis entity $c$ and a depression entity $d$ in the tweet, corresponding to two non-overlapping consecutive spans of length $k$ and $l$: $S_c=\{w_{c_1}, w_{c_2}, \ldots, w_{c_k}\}$ and  $S_d=\{w_{d_1}, w_{d_2}, \ldots, w_{d_l}\}$. In effect, this processing abstracts different lexical sequence in tweets to their meaning. 
\subsection{Entity Position-aware Module}
This module is designed to infuse the knowledge of the entity mention in basic neural models to effectively capture the contextual information w.r.t the entities.
The module consists of following three layers as:
\paragraph{\textbf{Position Embedding Layer:}}\label{position}
Inspired by the position encoding vectors used in \cite{collobert2011natural,zeng2014relation}, we define a position sequence relative to the cannabis entity $\{p_1^c, p_2^c, \ldots, p_l^c\}$, where
\begin{equation}
\footnotesize
  p_i^{c} =
    \begin{cases}
      i- c_1 &   i < c_1\\
      0 &   c_1 \leq i \leq c_k \\
      i-c_k & i> c_k
    \end{cases}       
\end{equation}
Here, $p_i^c$ is the relative distance of token $w_i$ to the cannabis entity and $c_1$ and $c_k$ are the beginning and end indices of the cannabis entity, respectively. In the same way, we computed the relative distance $p_i^d$ of token $w_i$ to the depression entity. This provides two position sequences $p^c=\{p_1^c, p_2^c, \ldots, p_n^c\}$ and $p^d=\{p_1^d, p_2^d, \ldots, p_n^d\}$. Later, for each position in the sequence, an embedding is learned with an embedding layer to producing two position embedding vectors, $P^c=\{P_1^c, P_2^c, \ldots, P_n^c\}$ for cannabis position embeddings and $P^d=\{P_1^d, P_2^d, \ldots, P_n^d\}$, both sharing a position embedding matrix P respectively. \\
\indent Further, we map each of the tokens from the input tweet $S$ to the pre-trained word embedding matrix $E \in \mathcal{R}^{V \times d}$ having the vocabulary size $V$ and dimension $d$. We used FastText\footnote{\url{https://bit.ly/36ldxJb}}, a pre-trained word embedding. We represent the input tweet after applying the word embedding as $e=\{e_1, e_2, \ldots e_n\}$, where $e_i \in \mathcal{R}^{d \times d}$.
Finally each word $i$ in the tweet $S$ is represented as the concatenation of the word embedding and relative distance of position embedding with respect to cannabis and depression:
\begin{equation}
\footnotesize
  x_i=e_i\oplus P_i^c \oplus P_i^d
\end{equation}
We denote the final representation of tweet as $x=\{x_1, x_2, \ldots x_n\}$.
The word feature and position feature representations compose a position-aware representation. 
\paragraph{ \textbf{Convolution Layer:}}
A combined representation of word and position embedding sequence $x$  is passed to the convolution layer, where filter $\textbf{F}\in \mathcal{R}^{m \times d}$ is convoluted over the context window  of $m$ words for each tweet. In order to ensure that the output of the convolution layer is of the same length as input, we performed the necessary zero-padding on the input sequence $x$. We call the zero-padded input as $\overline{x}$. 
\begin{equation}
\footnotesize
f_i^{m}=tanh(\textbf{F}.\overline{x}_{i:i+m-1} + b)
\end{equation}
where ${tanh}$ is the non-linear activation function and $b$ is a bias term. The feature map $f$ is generated by applying a given filter \textbf{F} to each possible window of words in a tweet, Mathematically,
\begin{equation}
\footnotesize
f^m = [f_{1}^{m}, f_{2}^{m}, \ldots,  f_{n}^{m}]
\end{equation}
We apply different length of context window $m \in M$, where $M$ is the set of context window length. Finally, we generate the hidden state $h_i$ at time $i$ as the concatenation of all the convoluted features by applying a different window size at time $i$.


\paragraph{\textbf{Entity Position-aware Attention Layer:}}
The intuition behind adding entity position-aware attention layer
is to select relevant contexts over irrelevant ones \cite{zhang2017position}.
This position-aware representation of entities in a tweet is further modulated by an ontology developed by domain experts. This enhancement enables us to selectively model attention and weigh entities in a tweet.
The position-aware attention layer takes as an input ${h_1, h_2, h_3,.....h_n}$ from the encoding module. We formulate an aggregate vector \textbf{q} mathematically as follows:
\begin{equation}
\footnotesize
    \textbf{q}= \frac{1}{n}\sum_{i=1}^{n}h_i
\end{equation}
The vector \textbf{q}, thus, stores the global, semantic, and syntactic information contained in a tweet. With the aggregate vector, we compute attention weight $a_i$ for each hidden state $h_i$ as 
\begin{equation}
\footnotesize
u_i = v^T\tanh(W_h h_i + W_q q+ W_c P_i^c + W_d P_i^d)
\end{equation}
\begin{equation}
\footnotesize
    \alpha_i = \frac{exp(u_i)}{\sum_{j=1}^{n}exp(u_j)}
\end{equation}
where, $ W_h, W_q \in R^{d_a \times d_h}; W_c, W_d \in R^{d_a \times d_p}; V \in R^{d_a} $
are parameters of the network, where $d_h$ is the dimension of the hidden states, $d_p$ is the dimension of position embedding, $d_a$ is the size of attention vector.
After applying the attention, the final tweet representation ${r}$ is computed as 
\vspace{-1.0 em}
\begin{equation}
\footnotesize
    R = \sum_{j=1}^{n}\alpha_j h_j
\end{equation}
\subsection{Encoding Module}
In the encoding module, we aim to obtain the semantic and task-specific contextualized representation of the tweet. We leverage the joint representation through BERT language representation model and Entity position-aware module. 
\begin{table*}[]
\centering
\resizebox{\linewidth}{!}{
\begin{tabular}{c|c|ccc}
\hline
\multirow{2}{*}{\textbf{Models}} & \multirow{2}{*}{\textbf{Techniques Used}} &
\multicolumn{3}{c}{\textbf{Cannabis-Depression RE}}  \\ \cline{3-5} 
 &  & \textbf{Precision} & \textbf{Recall} & \textbf{F\textsubscript{1}-Score} \\ \hline
 \hline
Baseline 1 & \begin{tabular}[c]{@{}c@{}}BERT\end{tabular} &64.49	&63.22	&63.85  \\ \hline
Baseline 2 & BioBERT &63.97	&62.15	&63.06  \\ \hline
Baseline 3 & BERT\textsubscript{PE} &60.64	&56.51	&58.50   \\ \hline
Baseline 4 & BERT\textsubscript{PE+PA} &65.41 	&65.25	& 64.50  \\ \hline
\textbf{Proposed Approach} & \textbf{Gated-K-BERT} &\textbf{66.41} &	\textbf{67.10} &   \textbf{	66.75} \\ \hline
\end{tabular}
}
\caption{Performance comparison our proposed model with the baselines methods.}
\label{results}
\end{table*}

Owing to its effective word and sentence level representation, BERT provide a task-agnostic architecture that has achieved state-of-the-art status for various NLP tasks \cite{maillard2019jointly,hewitt2019structural}. We use the pre-trained BERT model\footnote{\url{shorturl.at/nDJPY}} having $12$ Transformer layers ($L$), each having $12$ heads for self-attention and hidden dimension $768$. The input to the BERT model is the tweet $S=\{w_1, w_2, \ldots, w_n\}$. It returns the hidden state representation of each Transformer layer. Formally,
\begin{equation}
\footnotesize
    H_b^1, H_b^2, \ldots, H_b^L= BERT([w_1, w_2, \ldots, w_n])
\end{equation}
where, $H_i\in \mathbb{R}^{n \times h_b}$ and $h_b$ is the dimension of the hidden state representation obtained from BERT. We masked the representation of  \texttt{[CLS]} and \texttt{[SEP]} tokens with zero. We obtained the tweet representation via BERT model as follows:
\begin{equation}
\footnotesize
    B= \frac{1}{n}\sum_{j=1}^{n} H_b^{L-1}[j,:]
\end{equation}
In our experiments, the representation obtained from the second last ($L-1$) Transformer layer achieved the best performance on the task. The representation obtained from the last Transformer layer is  too close to the target functions (i.e., masked language model and next sentence prediction tasks) during pre-training of BERT, therefore may be biased to those targets. We also experiment with the \texttt{[CLS]} token representation obtained from BERT but that could not perform well in our experimental setting.
\subsubsection{Gated Feature Fusion}
The feature generated from CNN and BERT capture different aspect from the data. These features need to be used carefully to make most out of them. The joint feature obtained from concatenation or other arithmetic operations (sum, difference, min, max etc) often results in the poor joint representation. To mitigate this issue, we propose a gated feature fusion technique, which learn the most optimal way to join both the feature representation using a neural gate. This gate learn what information from CNN or BERT feature representation to keep or exclude during the network training. The gating behaviour is obtained through a \textit{sigmoid} activation which range between $0$ and $1$. We learn the joint representation $F$ using the gated fusion as follows: 
\begin{equation}
    \begin{split}
       h_R &= tanh(W_R.R) \\
       h_B &= tanh(W_B.B) \\
       g & = sigmoid(W_g.[R \oplus B])\\
       F &= g* h_R + (1-g)* h_B
    \end{split}
\end{equation}
where, $W_R, W_B$ and $W_g$ are the parameters.
Finally, the joint feature representation $F$ fed into a single layer feed-forward network with \textit{softmax} function to classify the tweet into one of the relation classes, $Y= \{$`\textit{reason}', `\textit{effect}', `\textit{addicted}', `\textit{ambiguous}'$\}$. More, formally,
\begin{equation}
    p(\hat{y}|S) = \text{softmax}(W.F+a)
\end{equation}
where $\hat{y} \in Y$, $W$ is a weight matrix and $a$ is the bias.
\section{Experimental Setup and Results}
Here, we present results\footnote{Hyper-parameter setting can be found in Appendix} on the cannabis-depression RE task. Thereafter, we will provide technical interpretation of the results followed by domain interpretation of the results.
\begin{table}[]
\centering
\resizebox{0.95\linewidth}{!}{
\begin{tabular}{l|l|l|l}
\hline
\textbf{Model} & \textbf{Precision} & \textbf{Recall} & \textbf{F\textsubscript{1}-Score} \\ \hline
\textbf{\begin{tabular}[c]{@{}l@{}}Proposed Model\\ Gated(CNN+PE+PA+BERT)\end{tabular}} & 66.41 & 67.10 & 66.75 \\ \hline
\textbf{-BERT} & 65.54 (0.87$\downarrow$) & 61.83 (5.27$\downarrow$ ) & 63.59 (3.16$\downarrow$) \\ \hline
\textbf{-Position-aware Attention} & 64.94 (1.47$\downarrow$) & 64.63 (2.47$\downarrow$) & 64.79 (1.96$\downarrow$) \\ \hline
\textbf{-Position Embedding} & 65.68 (0.73$\downarrow$) & 65.18(1.92$\downarrow$) & 65.43 (1.32$\downarrow$) \\ \hline
\textbf{-CNN} &60.55 (5.86$\downarrow$)  &57.26(9.84$\downarrow$)  &58.86 (7.89$\downarrow$)  \\ \hline
\end{tabular}
}
\caption{Ablation Study: the value within the bracket shows the absolute decrements in the model by removing the respective component. }
\label{ablation}
\end{table}

\begin{table}
\centering
\resizebox{0.75\linewidth}{!}{
\begin{tabular}{l|l|l|l}
\hline
\textbf{Model} & \textbf{Precision} & \textbf{Recall} & \textbf{F\textsubscript{1}-Score} \\ \hline
\textbf{\begin{tabular}[c]{@{}l@{}}Proposed Model\\ (-) Gated Fusion\end{tabular}} & 67.35 & 64.07 & 65.67 \\ \hline
\textbf{\begin{tabular}[c]{@{}l@{}}Proposed Model\\ (+) Gated Fusion\end{tabular}} & 66.41 & 67.10 & 66.75 \\ \hline
\end{tabular}
}
\caption{Performance comparison of our proposed model with/without gated fusion mechanism.}
\label{gated}
\end{table}
\subsection{Results}
The dataset utilized in our experiment is described in Section-\ref{dataset}. We used Recall, Precision and F\textsubscript{1}-Score to evaluate our proposed task against state-of-the-art relation extractor. As a baseline model, we used \textbf{\textit{BERT}}, \textbf{\textit{BioBERT}} and its various variation such as:\\
\textbf{BERT\textsubscript{PE}:} We extend the BERT with the position information (relative distance of the current word w.r.t cannabis/depression entities) obtained through ontology, as a position embedding along with the BERT embedding.\\
\textbf{BERT\textsubscript{PE+PA}:} We introduced additional component to the BERT\textsubscript{PE} model by deploying position-aware attention mechanism. \\
\indent Table-\ref{results} summarizes the performance of our model over the baselines.
Our proposed model significantly outperforms the state-of-the-art baselines on all the evaluation metrics. In comparison with the BERT \& BioBERT, our model achieves the absolute improvement of 2.9\% \& 3.69\% F\textsubscript{1}-Score respectively. Second, the results shows that infusing entity knowledge in the form of entity position-aware encoding with attention can assist in better relation classification. \\
\indent Among all the BERT-based approaches, we found that BERT\textsubscript{PE} did not perform well. Thus merely including position-aware encoding in the BERT framework does not help model to capture the entities information. This may be due to the inbuilt position embedding layer in the BERT model which treats the explicit position encoding as a noise. Further, our observation shows that BioBERT did not generalize well for our task in comparison to the BERT with minor reduction of $0.79$\% absolute F\textsubscript{1}-Score. Although BioBERT is trained on huge corpus of biomedical literature (PubMed \& PMC), however data being noisy hampered to performance.\\
\indent Interestingly, adding the entity position information in the form of the attention (BERT\textsubscript{PE+PA}) boosted the model performance. We report the performance absolute improvements of $0.92$\%, $2.03$\%, and $0.65$\%  Precision, Recall, and F\textsubscript{1}-Score points in comparison to the BERT model. This shows that position encoding and position attention when used collectively can assist in capturing complementary features.
Our final analysis reveals that solely concatenating two representation (CNN+BERT) may not be enough to capture how much information is required from both of these representations. Our method, which introduces the gated fusion mechanism can address this problem  as validated by the improved F\textsubscript{1}-Score (c.f. Table-\ref{gated}).
\subsection{Ablation Study}
To analyze the impact of various component of our model, we perform the ablation study (c.f. Table-\ref{ablation}) by removing one component from the proposed model and evaluate the performance. Results show that excluding BERT from the model significantly drop the recall of the model by $5.27$\%, and F\textsubscript{1}-Score by $3.16$\%. This shows that contextualized representation is highly necessary for the cannabis-depression classification task. \\
\indent We further observed that entity position-aware attention is highly crucial for improving the precision of the model. We report a reduction of $1.47$\% in terms of precision after excluding the position attention as the model component.\\
\indent Similarly, removing the position encoding from the input layer also lead to a reduced performance. While, excluding convolution layer from the model leads to significant drop in precision, recall, and F\textsubscript{1}-Score by $5.56$\%, $9.84$\%, and $7.89$\% respectively. Thus, we show that every component in the model is beneficial for the cannabis-depression relation extraction task.
\begin{table}
\begin{tabular}{l|l|l|l}
\hline
\textbf{Proposed Model} & \textbf{Precision} & \textbf{Recall} & \textbf{F\textsubscript{1}-Score} \\ \hline
\textbf{\begin{tabular}[c]{@{}l@{}} Entity Position-aware Attention\end{tabular}} & 66.41 & 67.10 & 66.75 \\ \hline
\textbf{\begin{tabular}[c]{@{}l@{}} Vanilla Attention\end{tabular}} & 66.43 & 64.91 & 64.30 \\ \hline
\end{tabular}
\caption{Performance comparison of our proposed model with position-aware attention over vanilla attention..}
\label{gated-1}
\end{table}
\begin{figure*}[h!]
\centering
\includegraphics[width=\linewidth]{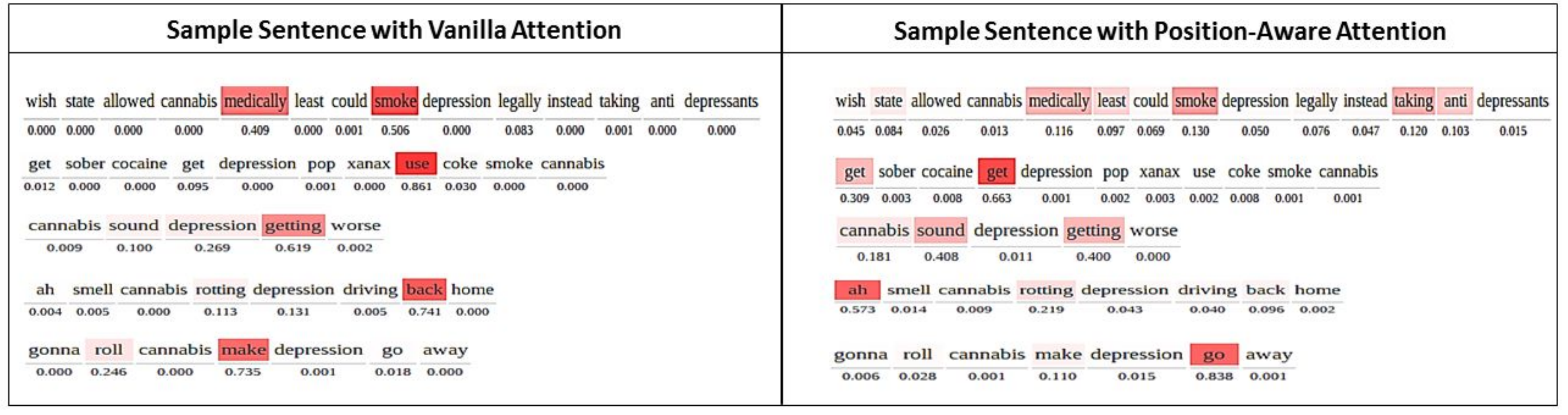}
\caption{\footnotesize Visualization of the vanilla attention (left) and position-aware attention (right). The actual label for the sentence 1, 2, 3, 4 and 5 are Reason, Reason, Ambiguous, Reason and Reason respectively. Vanilla attention incorrectly predicted it as Ambiguous, Ambiguous, Effect, Ambiguous and Cause for the sentence 1, 2, 3, 4 and 5 respectively. Our proposed model correctly predicted all the labels.}
\label{fig:Position attention}
\end{figure*}
\section{Discussion and Analysis}
Figure-\ref{fig:Position attention} shows
the visualization of attention weights assigned by
our model and the vanilla attention model. We find that the position-aware attention model learns to pay more attention to words that are informative for the relation. We also observe that the model tends to put more weight into cannabis/depression entities which are not observed in the case of the vanilla attention. 
For example, in the Sentence 1, (Figure-\ref{fig:Position attention}), the actual class label was `Reason', which the vanilla attention incorrectly predicted as `Ambiguous'. Instead of weighting just `medically' and `smoke', our proposed entity position-aware attention model distributes the weights across all the words including the cannabis and depression term.
The same can be observed for the all the other examples shown in Figure-\ref{fig:Position attention}. Experimental results of position-aware attention over vanilla attention is available in Table-\ref{gated-1}. 
\subsection{Domain-Specific Analysis}
To assess the performance on our model, we examined a set of correctly and incorrectly classified, tweets and came up with the following observations: 
\begin{itemize}
    \item \textbf{Correctly classified tweets generally contained clear relationship words:} For example, the following two tweets were correctly classified as expressing cannabis use to treat depression:\\
    ``\textit{weed really helps my depression so much ! i get less irritable, laugh, and so much more and people think it as the devil! f*** you mean}";\\ ``\textit{marijuana is seriously my best friend rn. it helps me sooo much with my depression and anxiety}."\\ Both tweet contained word ``help" that often times is used to convey a meaning indicating usage of a drug for the treatment of a certain condition.
    \item The following correctly classified example represented a case where relationship indicating ``treat" was expressed with a word ``for": \\
    ``\textit{I was forced to tell my family i have a medical for weed bc someone been ratting me out, try explaining medical marijuana for depression to a traditional thinking family, i wanna die}". \\
    \item Similarly, the following tweet were correctly classified as expressing situations where cannabis use is causing depression and/or making it worse: \\ ``\textit{me @ me when i realize weed is making me depressed but i keep smoking}". \\
    Both tweets contained clear relationship word expressing causation ``make/making".\\
    \item \textbf{The incorrectly classified tweets generally were more ambiguous and/or contained implied meanings.} For example, the following tweet was labeled as expressing ``cannabis use to treat depression" while our model classified it as ``ambiguous":\\ ``\textit{depression is hitting insufferable levels rn and hot damn i could use some weed.}" \\
    This is an example, where relationship is implied, and there are no clear relationship word expressed in the text.
    \item The same misclassification occurred with the following tweet: \\
    ``\textit{me: wow i think im depressed i should really go to therapy: doesnt do any of that and instead uses weed to increase the dopamine in my brain.}" \\
    In this case, the expression ``used weed to increase the dopamine…" implies use of marijuana to improve mood (in this cases depressive mood). \\
    Because DAO did not contain similar colloquial expressions to indicate depressive mood, our model failed to correctly classify this tweet.
\end{itemize}

\section{Conclusion}
This research explored a new dimension of social media in understanding the relationship between the cannabis use and depression. We introduced a state-of-the-art knowledge-aware attention framework that jointly leverages knowledge from the domain-specific DAO, DSM-5 in association with BERT for cannabis-depression RE task. Further, our result and domain analysis help us find associations of cannabis use with depression.
In order to establish a more accurate and precise Reason-Effect relationship between cannabis and depression from social media sources, our future study would take targeted user profiles in real-time and study the exposure of the user to cannabis over time informing public health policy.

\section*{Acknowledgments}
We acknowledge the support from National Institute on Drug Abuse (Grant No. 5R21DA044518-02) for conducting this research. All findings and opinions are of authors and not
sponsors.


%
%
%
\bibliography{plos}
\bibliographystyle{plos2015}

\end{document}